\documentclass[sigconf]{acmart}

\usepackage{algorithm}
\usepackage{algpseudocode}
\AtBeginDocument{%
  }

\setcopyright{acmlicensed}
\copyrightyear{2026}
\acmYear{2026}
\acmDOI{XXXXXXX.XXXXXXX}
\acmConference[ICONS'26]{International Conference on Neuromorphic Systems}{August 04--06,
  2026}{Chicago, Illinois}
\acmISBN{978-1-4503-XXXX-X/2026/08}

\begin{document}

\title{Sigma-Delta Neural Network Conversion on Loihi 2}


\author{Matthew Brehove}
\affiliation{%
  \institution{ChromoLogic LLC}
  \city{Monrovia}
  \state{California}
  \country{USA}
}
\email{mbrehove@chromologic.com}

\author{Sadia Anjum Tumpa}
\authornote{Corresponding author}
\affiliation{%
  \institution{Pennsylvania State University}
  \city{University Park}
  \state{Pennsylvania}
  \country{USA}
}
\email{sbt5360@psu.edu}

\author{Espoir Kyubwa}
\affiliation{%
  \institution{ChromoLogic LLC}
  \city{Monrovia}
  \state{California}
  \country{USA}
}
\email{ekyubwa@chromologic.com}

\author{Naresh Menon}
\affiliation{%
  \institution{ChromoLogic LLC}
  \city{Monrovia}
  \state{California}
  \country{USA}
}
\email{nmenon@chromologic.com}

\author{Vijaykrishnan Narayanan}
\affiliation{%
  \institution{Pennsylvania State University}
  \city{University Park}
  \state{Pennsylvania}
  \country{USA}
}
\email{vxn9@psu.edu}
\renewcommand{\shortauthors}{Brehove et al.}

\begin{abstract}
Neuromorphic computing aims to improve the efficiency of artificial neural networks by taking inspiration from biological neurons and leveraging temporal sparsity, spatial sparsity, and compute near/in memory. Although these approaches have shown efficiency gains, training these spiking neural networks (SNN) remains difficult. 
The original attempts at converting trained conventional analog neural networks (ANN) to SNNs used the rate of binary spikes to represent neuron activations. This required many simulation time steps per inference, which degraded efficiency. 
Intel's Loihi 2 is a neuromorphic platform that supports graded spikes which can be used to represent changes in neuron activation. 
In this work, we use Loihi 2's graded spikes to develop a method for converting ANN networks to spiking networks, which exploits temporal and spatial sparsity. 
We evaluated the performance of this network on Loihi 2 and compared it to NVIDIA's Jetson Xavier edge AI platform.
The results show that neuromorphic approaches achieve significant improvements in efficiency and latency (energy-delay product) over existing solutions.

\end{abstract}




\maketitle

\section{Introduction}
The increasing demand for more energy-efficient artificial intelligence systems has spurred the development of neuromorphic computing, an approach inspired by the structure and function of biological neurons. This field aims to enhance computational efficiency by leveraging principles such as temporal sparsity, spatial sparsity, and integrating compute capabilities near or within memory units. Unlike conventional AI computing architectures, which use globally accessible memory to perform dense matrix operations, neuromorphic systems keep compute near memory and process inputs as a stream of sparse, discrete spikes routed between neurons. These spiking neural networks (SNN) can reduce unnecessary computations by only transmitting spikes when there are significant changes to their input. SNNs also minimize energy use by only communicating between neurons selectively and by keeping memory close to computation units, reducing data movement and latency.

One notable hardware realization of these principles is Loihi\cite{loihi}, Intel’s neuromorphic processor for low-power, event-driven computations. 
The Loihi 2 \cite{loihi2} chip used in this work comprises 120 neuromorphic cores with each core supporting  $\sim$8k neurons. Each neuromorphic core holds its associated neuron's weights and internal state in local SRAM, enabling low-latency data storage. Loihi chips can be networked on a board to pass spikes between them and execute larger networks. 
Loihi 2's neurons can execute custom microcode, allowing researchers to implement specialized neuron types tailored to specific applications. Each Loihi 2 chip is 31 mm² and is deployed on either the single-chip ``Oheo Gulch" board, the 8-chip ``Kapoho Point" board or the 16-chip VPX board which is used in these experiments. Built on the Intel 4 process, Loihi 2 use much less than 1 Watt per chip. 

\begin{figure*}[!ht]
    \centering
   \includegraphics[width=0.9\linewidth]{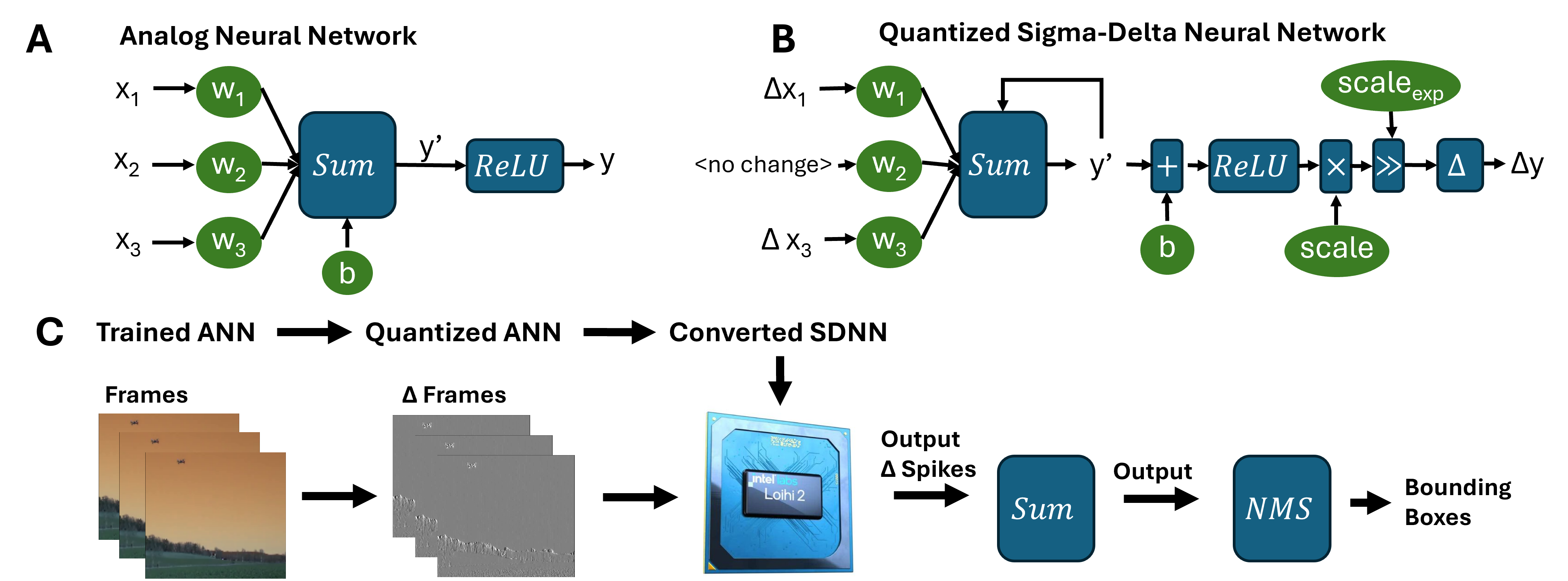}
        \caption{(A) In an ANN neuron all of the activations from the previous layer are multiplied by their respective weights to calculate the preactivation ($y'$). The ReLU operation is applied to calculate the activation ($y$) which is sent to the next layer.      
        (B) For a quantized sigma-delta neuron running on Loihi, the changes in the previous layer activations are multiplied by quantized weights and then added to a running sum to find $y'$. 
        It is then passed through a ReLU operation and rescaled by multiplication and right-shift. If this rescaled value has changed by more than a set threshold, a spike is generated with the value of that change which is represented as $\Delta y$.
        This approach allows for the neuron to only perform calculation when the input of the neuron has changed, but otherwise behaves exactly like a conventional ANN neuron. 
        (C) Overall workflow for our experiments. The quantized and converted network runs as an SDNN on Loihi 2. Frames are converted into sparse delta spikes, passed to Loihi, and the output spikes added to a running sum and interpreted as bounding boxes.}
        \Description{A three-panel figure illustrating the proposed sigma-delta neural network framework. Panel (A) shows the computation performed by a conventional artificial neuron, where weighted inputs are accumulated, a ReLU activation is applied, and the resulting activation is forwarded to the next layer. Panel (B) shows the corresponding sigma-delta neuron implementation on Loihi 2, where changes in the inputs are accumulated, passed through a ReLU operation, rescaled, and compared with a reference value to determine whether a spike encoding the output change should be emitted. Panel (C) presents the end-to-end deployment pipeline, beginning with video frames, followed by frame differencing and spike encoding, execution of the converted network on Loihi 2, accumulation of output spikes, and generation of object detection bounding boxes. Arrows indicate the flow of data between processing stages.}
      \label{fig: ann-sdnn}  
\end{figure*}

Training SNNs with binary spikes to accuracies comparable to conventional deep analog neural networks (ANN) has proven challenging, in part because the loss must be back-propagated over time~\cite{lee_training_2016}. 
In our experiments, training a spiking network takes $\sim10\times$ the time as an ANN with the same architecture (Section~\ref{sec: SLAYER training comparisions}).
Attempts to convert trained ANNs to SNNs have largely used the rate of spikes to represent activation values \cite{rueckauer_conversion_2017}. 
This approach requires many timesteps per input to accurately estimate output spike rates. 
Loihi 2's support for quantized graded spikes enables converting an ANN to a quantized graded-spike SNN that takes advantage of temporal and spatial sparsity while being easy to train. 
Loihi's graded spikes can hold up to 16 bit signed integer payloads which can represent quantized activations. This approach was inspired by the sigma-delta neural network (SDNN) \cite{oconnor_sigma_2016}. SDNNs accept incoming spikes representing changes in the activation of the previous layer, apply learned weights, update their internal state, and then pass the change in their state to the next layer if it is larger than a given threshold. Their operation closely mirrors that of a conventional ANN neuron (Fig.~\ref{fig: ann-sdnn}).  By updating their state only when the input changes, SDNNs naturally exhibit temporal sparsity. Additionally, their selective activation of a limited subset of neurons ensures spatial sparsity, reducing power consumption, and improving resource efficiency.
Our contributions are summarized as follows:
\\
$\bullet$ We propose a novel method to convert trained artificial neural networks (ANNs) into Sigma-Delta 
Neural Networks (SDNNs).\\
$\bullet$ We deploy the converted network on Loihi 2 in the field
demonstrating the feasibility of running converted SDNNs on neuromorphic hardware.\\
$\bullet$ We benchmark the converted network running on Loihi 2 against the original ANN network running on a Jetson Xavier edge AI accelerator 
and show that neuromorphic approaches achieve substantial improvements in energy-delay product.


\section{Related Works}

O’Connor and Welling introduced SDNNs to reduce redundant computations by transmitting only changes in activation~\cite{oconnor_sigma_2016}. This temporal difference encoding allows for computational savings while maintaining network performance. Shrestha et al.~\cite{shrestha_efficient_2023} implemented an SDNN on Intel’s Loihi 2 and benchmarked its performance against an NVIDIA Jetson platform. Their work demonstrated the potential advantages of neuromorphic hardware for SDNNs by taking advantage of Loihi 2’s event-driven processing. However, their approach relied on SLAYER~\cite{shrestha_slayer_2018}, a training framework designed for binary spiking networks. SLAYER requires the entire time series to be stored in memory during training. We found that this makes SDNN training more difficult and slower than conventional ANN training. Yousefzadeh et al. developed a method for converting ANNs to SDNNs~\cite{yousefzadeh_conversion_2019}, but this method still relied on binary spikes and was not compatible with modern model quantization approaches. In contrast, our approach can convert ANN networks trained and quantized with conventional pytorch tools to SDNNs capable of being deployed on Intel's Loihi 2 neuromorphic hardware. 

\section{Methodology}
\subsection{Conversion and Implementation on 
Loihi 2}
Loihi 2's neurocores are only capable of integer arithmetic, so the network needs to be quantized. Instead of developing our own quantization approach, we aimed to make our ANN-SDNN conversion method compatible with standard quantization schemes. We started with PyTorch's post-training static quantization~\cite{Paszke_PyTorch_An_Imperative_2019} which converts the weights ($w$) and activations ($x$) from floating-point (FP32) to 8-bit integer ($w_q$, $x_q$), while biases ($b$) were maintained in FP32. The quantization used an affine mapping defined by a floating-point scale ($s$) and an integer zero-point ($z$) for each tensor $x_{float}=s\times(x_q-z)$ for activations and similarly for weights $w_{float}=w\times(w_q-z)$. In PyTorch, the matrix multiplication is performed in integer arithmetic and the resulting integer accumulator is rescaled and combined with the FP32 bias before being re-quantized for processing by the next layer. The effective floating-point computation performed is: $y_{float i,k}=k\sum (x_{q,ik}-z_x)\times(w_{q,jk}-z_w))\times(s_x \times s_w)+b_{float,j}$.
 
For deployment on Loihi 2, direct application of the floating-point scales is not possible. The version of the sigma-delta neuron microcode developed by Intel only allows rescaling by powers of two (bit-shifts). 
 To maintain compatibility with the PyTorch quantization scales, we needed to modify the sigma-delta neuron microcode. We extracted the per-layer input scale ($s_x$), weight scale ($s_w$), and output scale ($s_y$) from the quantized PyTorch model using custom python code. 
 The necessary rescaling factor for a layer, $R=s_y \times s_x \times s_w$, was then implemented in the Loihi 2
 microcode using fixed-point arithmetic. 
 Specifically, the 24-bit accumulated pre-activation value within the neuron was multiplied by a pre-computed 24-bit integer scaling factor ($scale$) derived from $R$, and subsequently right-shifted by a pre-computed exponent ($scale_{exp}$) to approximate the division by $s_y$. 
 The Loihi 2
 microcode then implemented $y=(accumulator \times scale)>>scale_{exp}$ to ensure the result remained within an 8-bit range for the next layer's input, preventing overflow in the next layer's 24-bit accumulator. 
 The bias was similarly rescaled during conversion and added to the accumulator before scaling. This method was inspired by Jacob et al. \cite{jacob_quantization_2017}. 
 The full redesigned microcode is shown in pseudocode in Algorithm \ref{alg:new}.
Our method deploys the converted model on Loihi 2
using an initial delta layer executing on the CPU to produce graded delta spikes from a sequence of RBG frames [0, 255] where $spike_i = frame_i - frame_{i-1}$. 
The spikes are passed to Loihi 2 which performed the convolution operation and the sigma-delta neurons. 
The resulting spikes are received from Loihi 2 
and added to a running sum to calculate output activations. 
These output activation frames were then dequantized using the final layer scaling factor from the quantized PyTorch model. 
These floating point outputs are interpreted as a YOLOv3-style grid of $14\times14$ cells each with associated x-y locations and probability scores for each of the 3 bounding boxes (Fig.~\ref{fig: ann-sdnn}C). 
In our work, we leveraged nxkernel (v0.3.0) to deploy the model on Loihi 2, enabling more efficient mapping of our architecture to the hardware and better utilization of its event-driven computation capabilities. Inference was run in \textit{fall-through mode} which runs 16 SNN simulation timesteps for each input and thereby allowing spikes to propagate through the network and corresponding output.
This contrasts with nxkernel's \textit{pipelined mode} in which a new frame enters the network at each timestep and each layer of the network processes a different frame each step. Selecting \textit{fall-through mode} gives us a dramatic advantage in latency for a small cost in throughput. 


    
    
    
    
    
        
    

\begin{algorithm}
\caption{New Microcode Logic}
\label{alg:new}
\begin{algorithmic}[1]
\footnotesize
\State $y' \gets y' + \Delta x \cdot w_q$
\Comment{Accumulate input delta}

\State $y \gets \mathrm{ReLU}(y' + bias)$

\State $y \gets (y \times scale) \gg scale_{exp}$

\State $\Delta y \gets y - y_{ref}$

\If{$|\Delta y| > v_{th}$}
    \State Emit Spike($\Delta y$)
    \State $y_{ref} \gets y$
\EndIf
\end{algorithmic}
\end{algorithm}

\subsection{Dataset}
We chose drone-detection in ground-based video as our first test application for our ANN-SDNN conversion pipeline because these videos have high temporal sparsity, with few moving objects. 
Drone detection is also a highly relevant application for this context because low-power video inference could enable low-power persistent sensors to watch the sky for enemy drones.
We used the multi-sensor drone detection dataset~\cite{samaras_deep_2019} which includes 285 RGB videos, which have a resolution of 640$\times$512. 
We resized all input frames to fit the network input size of 448$\times$448. 
The dataset features four manually labeled classes of flying objects: drones, airplanes, helicopters, and birds. 
The sensor-to-target distance for drones ranged from 20 to 200 meters, and all video acquisitions were performed under daylight conditions. 
The videos were divided into training and test sets with 20$\%$ reserved for testing.

In order to directly compare ANN-SDNN converted networks with SDNNs that had been directly trained with SLAYER, we used the Driving dataset ~\cite{bojarski2017explainingdeepneuralnetwork,bojarski2016endendlearningselfdriving} which was used by Intel as a demonstration of SLAYER and Loihi 2's ability to run efficient SDNNs \cite{shrestha_efficient_2023}. 
This dataset is used to predict the steering angle of a vehicle based on the input from a dashboard RGB camera. 
It consists of images from a front-facing camera in a data collection vehicle coupled with the time-synchronized steering angle recorded from a human driver.
All input frames were resized to match the PilotNet network’s required input resolution of 66$\times$200$\times$3.

\begin{table}[t]
    \centering
    \caption{PilotNet and YOLO-KP model summary}
    \vspace{-4mm}
    \setlength{\tabcolsep}{4pt}
        \begin{tabular}{lrr}
        \toprule
                        & \textbf{PilotNet} & \textbf{YOLO-KP}\\ 
            \hline
            Type        & Conv/FC   & Conv \\
            Application & Angle Pred.& Obj. Detection      \\
            Parameters  & 0.35M     & 3.4M \\
            Neurons     & 12K    &  2M  \\
            Loihi Chips       & 1     & 5 \\
        \bottomrule
        \end{tabular}

    \vspace{-4mm}
    \label{tab:model_comparison}
    
\end{table}

\subsection{Model and Training}
 The model architecture used for drone detection (YOLO-KP) was developed by Intel and based on the YOLO v3-Tiny architecture~\cite{YOLO_v3_tiny}. 
 It was adapted to accommodate 
 the memory and architectural constraints of Loihi 2.
 The most important consideration for Loihi 2 is minimizing the number of neurons.
 Each Loihi 2 core can hold only a fixed number of neurons as each requires designated dendrite and axon resources. 
 Therefore, higher resolution and deeper networks require more Loihi 2 chips which in turn consume more power. 
 For this reason, max-pooling layers were replaced by strided convolutions for spatial down-sampling. 
 Residual layers were similarly removed to reduce memory requirements and eliminate the need to synchronize spikes arriving from different layers. 
 These removals greatly simplified the implementation. 
 The model in the lava-nc/lava-dl github repository~\cite{lava-dl} has been used as the basis for other works benchmarking video object detection on Loihi~\cite{barnell}. 
The PilotNet model used for the Driving dataset had an identical architecture to Intel's SLAYER trained model~\cite{shrestha_efficient_2023} and was small enough that it was able to run on a single Loihi chip. 
 Model summaries are shown in Table \ref{tab:model_comparison}.
 
 In order to demonstrate the flexibility of our ANN-SDNN conversion approach, we trained the ANN model using student-teacher distillation. We first trained a MobileNetV2 \cite{mobilenet} based teacher model on the same dataset. We then employed a masked L2 feature matching loss to align the penultimate layer representations between the teacher (MobileNetV2) and student (YOLO-KP) models.
 This feature distillation loss was combined with the standard detection loss using a weighting coefficient $\alpha$ = 0.069, allowing the student network to simultaneously optimize for detection performance while mimicking the richer feature representations learned by the teacher model.   
 The model hyper-parameters were tuned with Optuna.
 We used the albumentations package for data augmentation, Adam optimizer, Reduce-on-plateau learning rate scheduler and trained for 30 epochs.
The model was quantized to 8-bit per-tensor using PyTorch with HistogramObserver (activations) and MinMaxObserver (weights).

\begin{figure*}[!ht]
\centering
\includegraphics[width=0.85\linewidth]
   {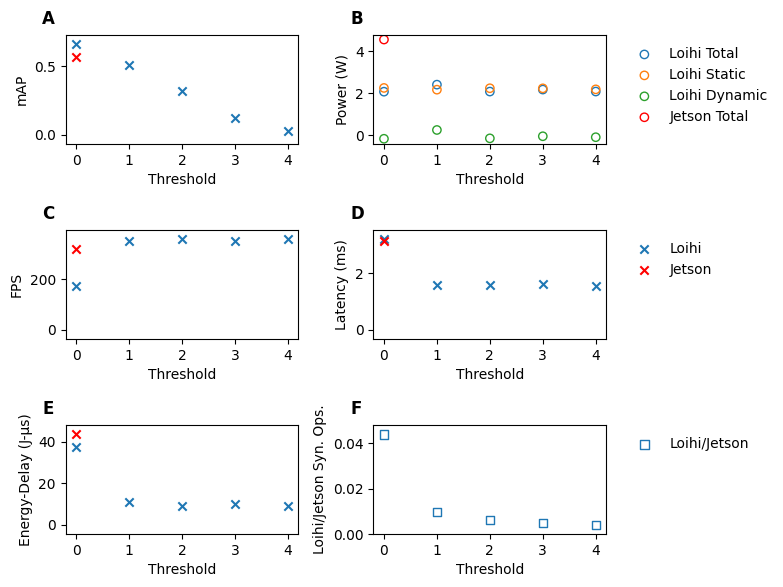}
      \caption{Results from the 
      YOLO-KP model as run on the Jetson Xavier in full precision and as a SDNN on Loihi 2 
      across threshold 0 to 4.
      (A) Mean average precision. (B) Power usage. (C) Frames per second. (D) Time between input and corresponding output. (E) Energy-delay product (J-$\mu s$) (F) Ratio of the Loihi 2 synaptic operations per frame and the number of multiply accumulate operations performed per frame in the ANN network on the Jetson Xavier. As expected, increasing the thresholds dropped the accuracy, increased the speed, and reduced the number of operations performed.}
      \Description{A six-panel figure comparing the performance of the full-precision YOLO-KP model running on a Jetson Xavier with the converted sigma-delta neural network running on Loihi 2 across output spike thresholds ranging from 0 to 4. The plots show how increasing the threshold affects detection accuracy, power consumption, throughput, latency, energy-delay product, and computational workload. Detection accuracy decreases as the threshold increases, while throughput improves, latency decreases, and the number of synaptic operations performed by Loihi 2 is reduced relative to the multiply-accumulate operations required by the original ANN. Power consumption remains substantially lower for Loihi 2 than for the Jetson Xavier across all threshold settings.}
      \label{fig: results} 
\end{figure*}

\subsection{Accuracy-Sparsity Tradeoff Exploration}

\begin{figure*}[!ht]
\centering
   \includegraphics[width=0.85\linewidth]{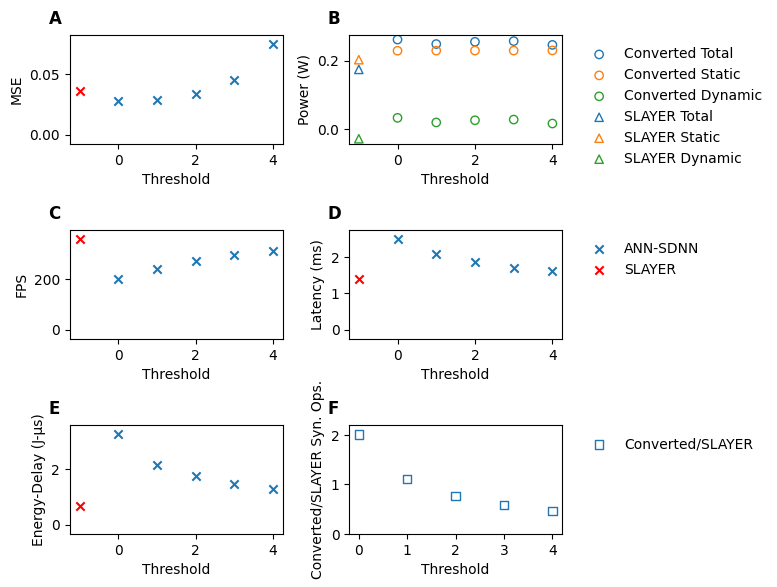}
      \caption{PilotNet results on Loihi 2 with varying thresholds of 0-4 for (A) Mean Square Error, (B) Power, (C) Frames per second, (D) 
      latency, (E) Energy-delay product, (F) Ratio of synaptic operations.}
      \Description{A six-panel figure showing the performance of the converted PilotNet model running on Loihi 2 as the output spike threshold is varied from 0 to 4. The plots illustrate the trade-offs between prediction accuracy, power consumption, throughput, latency, energy-delay product, and computational workload. Increasing the threshold reduces the number of synaptic operations, leading to higher throughput, lower latency, and improved energy efficiency, while gradually increasing the mean squared error. The figure demonstrates the balance between computational efficiency and prediction accuracy achieved through threshold selection.}
      \label{fig: results_pilotnet}
      
      \vspace{-3mm}
\end{figure*}

SDNNs control sparsity by tuning the threshold for propagating changes to the next layer.
When the thresholds are set to zero, the cumulative output exactly matches the quantized network. 
When thresholds are increased,
only larger changes are passed through the network. 
This decreases the number of spikes and the number of synaptic operations needed to execute the network while sacrificing accuracy that exists in the original quantized ANN. 
In order to explore this tradeoff, we raised the threshold from 0 to 4 for the converted YOLO-KP and PilotNet networks starting from the initial delta layer and the results are shown in Fig.~\ref{fig: results} and Fig.~\ref{fig: results_pilotnet} respectively.
This gave us a spectrum from the minimum sparsity and highest accuracy to higher sparsity and lower accuracy.

\subsection{Benchmarking Loihi 2}
Our experiments were run on an in-house Loihi VPX development board containing 16 
Loihi 2
chips and associated IO electronics and connected by 1 and 10 Gigabit Ethernet for control and data respectively. 
In order to measure the power and latency of the 
Loihi 2
board, we used 
the nxkernel API's
built in  profiling probes that report the power usage, timing, and core-level activity measurements. 
These probes measured static and total power.
Following the format of Shrestha et al.\cite{shrestha_efficient_2023}, we report static, dynamic, and total power. Static power is calculated only from the chips that were used by the network. 
The dynamic power was calculated as the measured total power minus measured static power. 
The reported total power is the sum of the adjusted static power and the dynamic power. 

Since integrated circuit design can generally trade speed for energy-efficiency, the energy-delay product (EDP) is used as a measure of overall efficiency of an integrated circuit \cite{laros_iii_energy_2013}. The EDP reported here is calculated as $power \times timestep \times latency$. 
Measurements of the accuracy of Loihi 2 were performed separately by running the entire test set through the Loihi 2 board, saving the output spikes, and then evaluating the accuracy after the run. In order to better understand how IO impacts the execution speed of Loihi 2 in our applications, we loaded 100 frames into on-chip memory and replayed these 10,000 times in order to measure power and latency in the absence of any IO bottlenecks (Fig.~\ref{fig: results_unbottlenecked}). This approach for reporting power, latency, and EDP is consistent with Intel's recommended reporting structure for Loihi benchmarks \cite{shrestha_efficient_2023}.

\subsection{Benchmarking NVIDIA Jetson}
To compare our results on Loihi with conventional approaches, we ran the same network as an 16-bit floating point ANN on an NVIDIA Jetson Xavier edge AI board using the TensorRT library. The tegrastats tool provided by NVIDIA was used for power measurement and the VDD\_CPU\_GPU\_CV and VDD\_SOC power channels were combined to estimate the total power used for network execution. It should be noted that tegrastats on Xavier was not capable of measuring the power used solely by the GPU. Therefore, the comparison between the Xavier and Loihi 2 is not an identical comparison since we report only the power used by the Loihi 2 chip but must report the power used by the whole system-on-chip (SOC) in the case of the Xavier. This choice was made to be consistent with Intel's comparisons to a similar Jetson device\cite{shrestha_efficient_2023}. 
To separate GPU execution from CPU preprocessing, initialization and preprocessing were performed first, then execution was paused and resumed using TensorRT. The network was run with a batch size of 1 inside a python for-loop without the system GUI running. 
\section{Results}

\subsection{Model Accuracy}
The MobileNetV2 teacher model achieved a mean-average precision (mAP) of 0.72, while our knowledge-distilled student model (YOLO-KP) reached a higher mAP of 0.74. This performance improvement is notable considering that YOLO-KP without knowledge distillation never exceeded 0.6 mAP in our experiments. The comparable parameter count between YOLO-KP and MobileNetV2 likely contributed to the student model's ability to effectively leverage the teacher's representations. After quantization, the accuracy decreased modestly to 0.62 mAP. These results were calculated across our complete test set of 58 videos containing 18,214 frames. For context, the authors of the multi-sensor drone dataset reported mAP values ranging from 0.58 to 0.84 depending on target distance when using a larger, full-precision model~\cite{samaras_deep_2019}.
For PilotNet, performance is evaluated using mean squared error (MSE) and the configuration with threshold 0 achieves the lowest MSE of 0.0278.

\begin{figure*}[!ht]
\centering
   \includegraphics[width=0.8\linewidth]{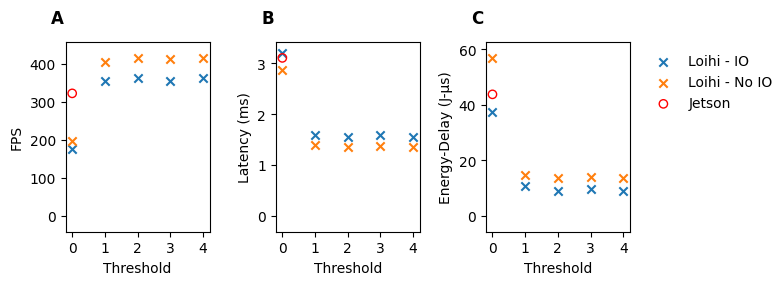}
      \caption{Results from the YOLO-KP model as run on Loihi 2 with and without off-board IO with varying thresholds from 0 to 4 to adjust sparsity. Input spikes were loaded into Loihi memory so that the model could run without off-board communication.}
      \label{fig: results_unbottlenecked}
      \Description{A multi-panel figure comparing the performance of the YOLO-KP model on Loihi 2 with and without off-board input/output communication as the output spike threshold is varied from 0 to 4. The plots show that removing off-board communication substantially improves throughput and reduces latency while maintaining similar trends in detection performance and energy efficiency across threshold values. Increasing the threshold reduces computational workload by decreasing the number of synaptic operations, resulting in higher processing speed and lower latency, with a corresponding trade-off in detection accuracy.}
\end{figure*}
\subsection{SLAYER Training Comparisons}
\label{sec: SLAYER training comparisions}
In order to compare the our ANN-SDNN conversion approach with Intel's SLAYER training algorithm, we trained the YOLO-KP SDNN network with the same architecture on the same Multi-Sensor Drone Detection Dataset. 
Training was conducted for 50 epochs with a batch size of 16, a starting learning rate of $2\times10^{-5}$ was used with a learning rate scheduler of factor 0.1. 
The results (Table~\ref{tab: slayer_vs_quantized_converted_model}) highlight significant improvements in both training efficiency and performance of our conversion approach. Our quantized model achieves higher mAP (0.62 vs 0.37) demonstrating better detection accuracy than SLAYER. Additionally, ANN training only took 5 minutes per epoch compared to 55 minutes for SLAYER on our NVIDIA GTX 4090. The conversion process only added ~30 seconds for post-training quantization.

\begin{table}[t]
    \centering
    \caption{Performance comparison of our training approach to Intel's SLAYER training approach ~\cite{shrestha_slayer_2018} in terms of training time and accuracy (mean average precision).}
    \vspace{-4mm}
    \setlength{\tabcolsep}{3pt}
    \begin{tabular}{ccccc}
    \hline \hline
    \textbf{Model}          & \begin{tabular}[c]{@{}c@{}} \textbf{Time per} \\ \textbf{Epoch (min) }\end{tabular} & \begin{tabular}[c]{@{}c@{}} \textbf{Completion} \\\textbf{Time (hrs)}\end{tabular} & \textbf{Memory} & \textbf{mAP}  \\ \hline \hline
    SLAYER    & 55    & 45.80 & 9.4G  & 0.37 \\ \hline
    Our Approach
    & 5    & 3.40 & 8.6G   & 0.62 \\ \hline \hline
    \end{tabular}
    \label{tab: slayer_vs_quantized_converted_model} 
    \vspace{-6mm}
    
\end{table}

\begin{table*}[ht]
    \centering
    \footnotesize
    \caption{
    Performance comparison of Loihi 2 and Jetson for YOLO-KP and PilotNet at zero threshold.
    $\downarrow$: lower is better, $\uparrow$: higher is better.}
    \vspace{-4mm}
    \resizebox{\textwidth}{!}{ 
        \begin{tabular}{llcccc}
        \toprule
            Model     & Platform      & Energy/frame (mJ $\downarrow$) & Latency (ms $\downarrow$) & Throughput (FPS $\uparrow$) & EDP ($\mu Js\downarrow$) \\ 
            \hline
            YOLO-KP SDNN   & Loihi          & 11.73                    & 3.20                       & 175.6                           & 37.59 \\
            YOLO-KP SDNN  & Loihi no IO    & 19.89                    & 2.86                       & 196.5                           & 56.93 \\
            YOLO-KP ANN  & Jetson Xavier  & 14.10                    & 3.11                       & 321.9                           & 43.80 \\
            \hline
            PilotNet SDNN  & Loihi          & 1.31                   &  2.50                     & 200.2                           & 3.26 \\
            PilotNet SDNN  & Loihi no IO    & 1.06                    &  1.13                       & 442.9                           & 1.20 \\
            PilotNet ANN~\cite{shrestha_efficient_2023}  & Jetson Nano    & 21.94                   &  5.77                     & 173.2                           & 126.20 \\            
            \hline
            \hline
                        & &  Energy Ratio($\downarrow$) & Latency Ratio($\downarrow$) & Throughput Ratio($\uparrow$) & EDP Ratio ($\downarrow$) \\ 
            \hline
            YOLO-KP          &    & 0.832 & 1.031 & 0.545 & 0.858 \\
            YOLO-KP (no IO) && 1.410 & 0.922 & 0.610 & 1.300 \\
            PilotNet          &    & 0.060 & 0.433 & 1.156 & 0.026 \\
            PilotNet (no IO)& & 0.0483 & 0.196 & 2.56 & 0.00948 \\

        \bottomrule
        
        \end{tabular}
    }
     
    \label{tab:loihi_jetson_comparison}
    
\end{table*}

\subsection{Power, Latency, and Throughput}
\label{sec: Power}

The results in Fig.~\ref{fig: results}, obtained from Loihi 2 activity probes show that the YOLO-KP SDNN had 0.056 of the synaptic operations as the ANN network and that, as expected, 
this ratio decreased as the threshold increased.
Increasing the sigma-delta threshold from 0 to 4 increased the sparsity from 18$\times$ to 250$\times$ which improved the throughput from 176 FPS to 362 FPS. 
The total power usage was lower for Loihi 2 compared to the Jetson and remained relatively consistent for all experiments as it is dominated by static draw. 
The model running on Loihi 2 executes a single time step in 3.20 ms and uses 11.73 mJ of energy.
The IO-constrained YOLO-KP SDNN on Loihi 2 achieves over 1.2$\times$ lower EDP.
The throughput and latency are somewhat worse than the Jetson Xavier which processes an entire frame in 3.1 ms and uses 14.1 mJ of energy (Fig.~\ref{fig: results}). 
For the unbottlenecked trials that avoided using IO, the speed increased to 2.86 ms per timestep and used 19.89 mJ of energy per inference (Fig.~\ref{fig: results_unbottlenecked}).
For PilotNet, the SDNN implementation on Loihi 2 demonstrates substantial efficiency improvements over the ANN baseline on Jetson Nano. 
The Loihi-based PilotNet SDNN model achieves significantly lower energy per frame (1.31 mJ vs. 21.94 mJ) and reduced latency (2.50 ms vs. 5.77 ms), resulting in a much lower energy-delay product (3.26 $\mu$Js vs. 126.20 $\mu$Js). 
When I/O overhead is removed, performance improves further, with energy dropping to 1.06 mJ, latency to 1.13 ms, and throughput increasing to 442.9 FPS, yielding an EDP of just 1.20 $\mu$Js. 
A detailed performance comparison of Loihi 2 and Jetson for YOLO-KP and PilotNet at zero threshold is provided in Table~\ref{tab:loihi_jetson_comparison}.
Overall, these results highlight the strong energy efficiency and latency advantages of the SDNN approach on Loihi 2 for PilotNet compared to conventional ANN deployment on edge GPUs.

\subsection{Comparison with Intel's published work}
Intel has previously published work comparing PilotNet SDNN run on Loihi 2 with the same network run on an NVIDA Jetson Orin Nano accelerator\cite{shrestha_efficient_2023}. In this study we re-ran Intel's trained PilotNet SDNN using the recently released nxkernel API and found that the execution time was substantially improved. In our study, the SDNN implementation achieves lower energy per frame (1.31 mJ) and latency (2.50 ms) which further improves in the no-I/O configuration to 1.06 mJ and 1.13 ms, yielding a strong reduction in energy-delay product (EDP) from 3.26 $\mu$Js to 1.20 $\mu$Js. Intel’s results show a comparable energy figure under standard operation (1.26 mJ)\cite{shrestha_efficient_2023}, but significantly higher end-to-end latency (65.4 ms)\cite{shrestha_efficient_2023}, suggesting substantial I/O or system overheads in their pipeline. These results demonstrate the efficiency benefits of Loihi-based SDNNs, while highlighting that system integration and I/O overheads play a critical role in determining end-to-end performance.

\subsection{Field Deployment at Silent Swarm 2025}
\begin{figure*}[!ht]
\centering
\includegraphics[width=0.8\linewidth]{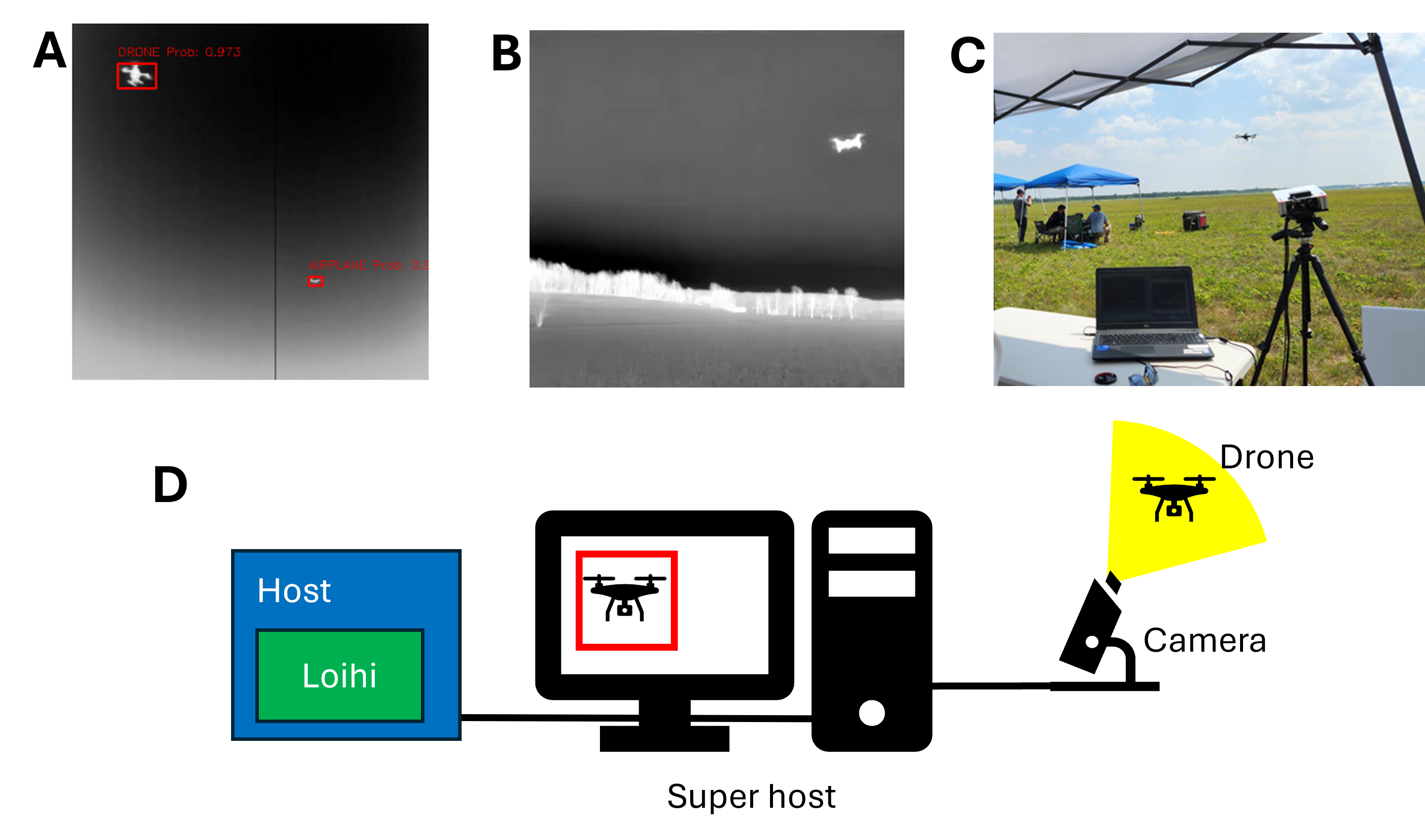}
      \caption{Field deployment of YOLO-KP SDNN at US Navy Silent Swarm 2025. (A) Sample IR frame with drone and airplane (SS25 dataset). (B) Representative MSDD training frame for comparison. (C) Field setup with IR camera and Loihi 2 VPX connected to host laptop.
      (D) System diagram: the host performs frame quantization and delta encoding, transmitting input spikes to Loihi 2 over Ethernet; output spikes are returned to the host for spike summation and non-maximum suppression before bounding boxes are overlaid on the live video feed. 
      }
      \label{fig: field_deployment} 
      \Description{A four-panel figure illustrating the real-world deployment and evaluation of the proposed sigma-delta neural network system. The first two panels show representative infrared images from the Silent Swarm 2025 field dataset and the MSDD training dataset, highlighting differences in scene appearance and target characteristics. The third panel is a photograph of the experimental setup, including a camera, a Loihi 2 VPX board, and a host laptop used during field testing. The fourth panel is a block diagram showing the processing pipeline, where infrared frames are quantized and converted into delta spikes on the host, processed by Loihi 2, and the resulting output spikes are accumulated and post-processed to produce object detection bounding boxes displayed on the video stream.}
\end{figure*}

To validate the feasibility of deploying the converted SDNN in an operational setting, the YOLO-KP network was deployed on the Loihi 2 VPX board at the US Navy's Silent Swarm 2025 (SS25) experimentation event (Fig.~\ref{fig: field_deployment}). While the YOLO-KP architecture has previously been demonstrated on Loihi 2 for object detection from an airborne platform \cite{barnell}, a stationary ground-based sensor is a more natural fit for neuromorphic processing, as only the target moves within the field of view. The network was trained as an ANN on the IR video subset of the MSDD dataset and run on Loihi 2 in fall-through mode. The deployed system operated in real time, achieving 112 FPS with a latency of 5.0 ms and an energy consumption of 20 mJ per frame. These results demonstrate that ANN-to-SDNN conversion produces systems capable of sustained, real-time neuromorphic inference under field conditions; a non-trivial result given the added complexity of live data ingestion, hardware integration, and environmental variability absent from controlled benchmarks.
Detection performance reflected the substantial domain gap between MSDD and SS25: a model trained on MSDD achieved 0.03 mAP on SS25 data, improving 10× to 0.30 mAP with minimal fine-tuning on a small SS25 subset. This rapid adaptation highlights the transferability of the ANN-to-SDNN conversion approach. The remaining performance gap reflects the presence of novel drone types, including vertical-takeoff-and-landing platforms, and greater background complexity not represented in the MSDD training set. The field conditions also introduced higher camera motion, resulting in reduced temporal sparsity (13\% average sparsity after delta encoding versus 2\% for MSDD), which accounts for the difference in efficiency metrics relative to the controlled benchmarks in Section~\ref{sec: Power}

\section{Discussion}

We were successfully able to train ANN-based models for video understanding and control tasks, quantize them, and deploy them as SDNNs on Loihi 2.
 Across both YOLO-KP and PilotNet, the deployed networks leveraged significant sparsity gains relative to their ANN counterparts e.g. $>18\times$ sparsity improvement for YOLO-KP compared to the original ANN (Fig.~\ref{fig: results}F). 
 For both YOLO-KP and PilotNet, the Loihi 2 SDNN implementation substantially reduces energy per frame and energy-delay product compared to conventional ANN deployment, with PilotNet achieving over 16$\times$ lower energy and $\sim 40\times $  lower EDP.
Loihi achieves competitive latency and throughput, with further improvements observed when I/O bottlenecks are removed and thresholds are increased.
In comparing the two platforms, a key design choice of Loihi is the use of on-chip SRAM for all weights and activations. 
This reduces memory access latency but increases static power consumption, which scales with the number of Loihi chips and makes higher-resolution inputs more power-intensive.

The PilotNet results also allow a direct comparison between our conversion approach and SLAYER-trained SDNNs. Because our method converts standard quantized ANNs to SDNNs, it is fully compatible with conventional PyTorch training pipelines, enabling the use of off-the-shelf optimizers, data augmentation, and quantization tools without modification. This flexibility comes with an accuracy advantage: our converted PilotNet achieves lower MSE than the SLAYER-trained equivalent as shown in Fig. \ref{fig: results_pilotnet}A, consistent with the YOLO-KP results. The converted network does exhibit somewhat lower sparsity than a SLAYER-trained model, since sparsity is not explicitly encouraged during training, but it nonetheless captures the majority of the efficiency benefits available from sparse neuromorphic execution.

Even if compilation and IO improvements are implemented to decrease the execution time of these networks on Loihi 2, traditional frame-based cameras may not fully leverage Loihi's asynchronous event-driven architecture. 
Event-based cameras, with their near-continuous temporal resolution and inherent sparsity, could provide a more natural pairing with neuromorphic hardware like Loihi 2, potentially unlocking greater efficiency advantages for video analysis tasks. 
This complementary hardware approach may explain why lower resolution videos have shown more favorable comparisons with conventional accelerators \cite{shrestha_efficient_2023}. Additionally, a larger single Loihi 2 chip could offer the key benefit of reducing costly inter-chip communications making spiking networks more attractive.

\begin{acks}
The authors thank Mark Barnell and Dr. Qing Wu (Air Force Research Laboratory) for providing the Loihi 2 board, and Intel’s Loihi team, especially Sumit Bam Shrestha, for their support. This work was supported by the U.S. Navy (N6833522C0487) and SRC/DARPA JUMP 2.0 PRISM Center.
\end{acks}



\bibliographystyle{ACM-Reference-Format}
\bibliography{sample-base}

@misc{shrestha_slayer_2018,
	title = {{SLAYER}: {Spike} {Layer} {Error} {Reassignment} in {Time}},
	shorttitle = {{SLAYER}},
	urldate = {2024-02-29},
	publisher = {arXiv},
	author = {Shrestha, Sumit Bam and Orchard, Garrick},
	month = sep,
	year = {2018},
	note = {arXiv:1810.08646 [cs, stat]},
}

@article{samaras_deep_2019,
	title = {Deep {Learning} on {Multi} {Sensor} {Data} for {Counter} {UAV} {Applications}—{A} {Systematic} {Review}},
	volume = {19},
	copyright = {https://creativecommons.org/licenses/by/4.0/},
	issn = {1424-8220},
	
	doi = {10.3390/s19224837},
	language = {en},
	number = {22},
	urldate = {2024-05-13},
	journal = {Sensors},
	author = {Samaras, Stamatios and Diamantidou, Eleni and Ataloglou, Dimitrios and Sakellariou, Nikos and Vafeiadis, Anastasios and Magoulianitis, Vasilis and Lalas, Antonios and Dimou, Anastasios and Zarpalas, Dimitrios and Votis, Konstantinos and Daras, Petros and Tzovaras, Dimitrios},
	month = nov,
	year = {2019},
	pages = {4837},
}

@article{rueckauer_conversion_2017,
	title = {Conversion of {Continuous}-{Valued} {Deep} {Networks} to {Efficient} {Event}-{Driven} {Networks} for {Image} {Classification}},
	volume = {11},
	issn = {1662-453X},

	doi = {10.3389/fnins.2017.00682},
	language = {English},
	urldate = {2024-06-26},
	journal = {Frontiers in Neuroscience},
	author = {Rueckauer, Bodo and Lungu, Iulia-Alexandra and Hu, Yuhuang and Pfeiffer, Michael and Liu, Shih-Chii},
	month = dec,
	year = {2017},
	note = {Publisher: Frontiers},
}

@misc{shrestha_efficient_2023,
	title = {Efficient {Video} and {Audio} processing with {Loihi} 2},
	urldate = {2024-10-17},
	publisher = {arXiv},
	author = {Shrestha, Sumit Bam and Timcheck, Jonathan and Frady, Paxon and Campos-Macias, Leobardo and Davies, Mike},
	month = oct,
	year = {2023},
	note = {arXiv:2310.03251},
}

@inproceedings{yousefzadeh_conversion_2019,
	address = {Hsinchu, Taiwan},
	title = {Conversion of {Synchronous} {Artificial} {Neural} {Network} to {Asynchronous} {Spiking} {Neural} {Network} using sigma-delta quantization},
	copyright = {https://ieeexplore.ieee.org/Xplorehelp/downloads/license-information/IEEE.html},
	isbn = {978-1-5386-7884-8},

	doi = {10.1109/AICAS.2019.8771624},
	language = {en},
	urldate = {2025-02-21},
	booktitle = {2019 {IEEE} {AICAS}},
	publisher = {IEEE},
	author = {Yousefzadeh, Amirreza and Hosseini, Sahar and Holanda, Priscila and Leroux, Sam and Werner, Thilo and Serrano-Gotarredona, Teresa and Barranco, Bernabe Linares and Dhoedt, Bart and Simoens, Pieter},
	month = mar,
	year = {2019},
	pages = {81--85},
}

@misc{oconnor_sigma_2016,
	title = {Sigma {Delta} {Quantized} {Networks}},

	doi = {10.48550/arXiv.1611.02024},
	urldate = {2025-02-21},
	publisher = {arXiv},
	author = {O'Connor, Peter and Welling, Max},
	month = nov,
	year = {2016},
	note = {arXiv:1611.02024 [cs]},
}

@misc{lee_training_2016,
	title = {Training {Deep} {Spiking} {Neural} {Networks} using {Backpropagation}},
	
	doi = {10.48550/arXiv.1608.08782},
	urldate = {2025-03-06},
	publisher = {arXiv},
	author = {Lee, Jun Haeng and Delbruck, Tobi and Pfeiffer, Michael},
	month = aug,
	year = {2016},
	note = {arXiv:1608.08782 [cs]},
}

@incollection{laros_iii_energy_2013,
	address = {London},
	title = {Energy {Delay} {Product}},
	isbn = {978-1-4471-4492-2},
	booktitle = {Energy-{Efficient} {High} {Performance} {Computing}: {Measurement} and {Tuning}},
	publisher = {Springer London},
	author = {Laros III, James H. and Pedretti, Kevin and Kelly, Suzanne M. and Shu, Wei and Ferreira, Kurt and Vandyke, John and Vaughan, Courtenay},
	year = {2013},
	doi = {10.1007/978-1-4471-4492-2_8},
	pages = {51--55},
}

@misc{jacob_quantization_2017,
	title = {Quantization and {Training} of {Neural} {Networks} for {Efficient} {Integer}-{Arithmetic}-{Only} {Inference}},
	
	doi = {10.48550/arXiv.1712.05877},
	language = {en},
	urldate = {2025-05-03},
	publisher = {arXiv},
	author = {Jacob, Benoit and Kligys, Skirmantas and Chen, Bo and Zhu, Menglong and Tang, Matthew and Howard, Andrew and Adam, Hartwig and Kalenichenko, Dmitry},
	month = dec,
	year = {2017},
	note = {arXiv:1712.05877 [cs]},
}

@inproceedings{Paszke_PyTorch_An_Imperative_2019,
author = {Paszke, Adam and Gross, Sam and Massa, Francisco and Lerer, Adam and Bradbury, James and Chanan, Gregory and Killeen, Trevor and Lin, Zeming and Gimelshein, Natalia and Antiga, Luca and Desmaison, Alban and Kopf, Andreas and Yang, Edward and DeVito, Zachary and Raison, Martin and Tejani, Alykhan and Chilamkurthy, Sasank and Steiner, Benoit and Fang, Lu and Bai, Junjie and Chintala, Soumith},
booktitle = {Advances in Neural Information Processing Systems 32},
editor = {Wallach, H. and Larochelle, H. and Beygelzimer, A. and d'Alché-Buc, F. and Fox, E. and Garnett, R.},
pages = {8024--8035},
publisher = {Curran Associates, Inc.},
title = {{PyTorch: An Imperative Style, High-Performance Deep Learning Library}},
url = {},
year = {2019}
}

@ARTICLE{loihi,
  author={Davies, Mike and Srinivasa, Narayan and Lin, Tsung-Han and Chinya, Gautham and Cao, Yongqiang and Choday, Sri Harsha and Dimou, Georgios and Joshi, Prasad and Imam, Nabil and Jain, Shweta and Liao, Yuyun and Lin, Chit-Kwan and Lines, Andrew and Liu, Ruokun and Mathaikutty, Deepak and McCoy, Steven and Paul, Arnab and Tse, Jonathan and Venkataramanan, Guruguhanathan and Weng, Yi-Hsin and Wild, Andreas and Yang, Yoonseok and Wang, Hong},
  journal={IEEE Micro}, 
  title={Loihi: A Neuromorphic Manycore Processor with On-Chip Learning}, 
  year={2018},
  volume={38},
  number={1},
  pages={82-99},
  doi={10.1109/MM.2018.112130359}}

@INPROCEEDINGS{loihi2,
  author={Orchard, Garrick and Frady, E. Paxon and Rubin, Daniel Ben Dayan and Sanborn, Sophia and Shrestha, Sumit Bam and Sommer, Friedrich T. and Davies, Mike},
  booktitle={IEEE SiPS 2021}, 
  title={Efficient Neuromorphic Signal Processing with Loihi 2}, 
  year={2021},
  volume={},
  number={},
  pages={254-259},
  doi={10.1109/SiPS52927.2021.00053}}

@INPROCEEDINGS{YOLO_v3_tiny,
  author={Adarsh, Pranav and Rathi, Pratibha and Kumar, Manoj},
  booktitle={ICACCS 2020}, 
  title={YOLO v3-Tiny: Object Detection and Recognition using one stage improved model}, 
  year={2020},
  volume={},
  number={},
  pages={687-694},
  doi={10.1109/ICACCS48705.2020.9074315}}

@misc{lava-dl,
	title = {lava-nc/lava-dl},
	copyright = {BSD-3-Clause},
	urldate = {2025-03-06},
	publisher = {Lava},
	month = jan,
	year = {2025},
	note = {original-date: 2021-09-29T15:32:51Z},
}

@INPROCEEDINGS{barnell,
  author={Barnell, Mark and Raymond, Courtney and Loomis, Lisa and Vidal, Francesca and Brown, Daniel and Isereau, Darrek},
  booktitle={2024 IEEE HPEC}, 
  title={Spike-Driven YOLO: Ultra Low-Power Object Detection with Neuromorphic Computing}, 
  year={2024},
  volume={},
  number={},
  pages={1-5},
  doi={10.1109/HPEC62836.2024.10938424}}

@misc{mobilenet,
	title = {{MobileNetV2}: {Inverted} {Residuals} and {Linear} {Bottlenecks}},
	shorttitle = {{MobileNetV2}},
	
	doi = {10.48550/arXiv.1801.04381},
	urldate = {2025-03-28},
	publisher = {arXiv},
	author = {Sandler, Mark and Howard, Andrew and Zhu, Menglong and Zhmoginov, Andrey and Chen, Liang-Chieh},
	month = mar,
	year = {2019},
	note = {arXiv:1801.04381 [cs]},
}

@misc{bojarski2017explainingdeepneuralnetwork,
      title={Explaining How a Deep Neural Network Trained with End-to-End Learning Steers a Car}, 
      author={Mariusz Bojarski and Philip Yeres and Anna Choromanska and Krzysztof Choromanski and Bernhard Firner and Lawrence Jackel and Urs Muller},
      year={2017},
      eprint={1704.07911},
      archivePrefix={arXiv},
      primaryClass={cs.CV},
      
}

@misc{bojarski2016endendlearningselfdriving,
      title={End to End Learning for Self-Driving Cars}, 
      author={Mariusz Bojarski and Davide Del Testa and Daniel Dworakowski and Bernhard Firner and Beat Flepp and Prasoon Goyal and Lawrence D. Jackel and Mathew Monfort and Urs Muller and Jiakai Zhang and Xin Zhang and Jake Zhao and Karol Zieba},
      year={2016},
      eprint={1604.07316},
      archivePrefix={arXiv},
      primaryClass={cs.CV},
      url={https://arxiv.org/abs/1604.07316}, 
}

\end{document}